\documentclass{article}

\usepackage{amsmath,amsfonts,bm}
\usepackage{amssymb, amsthm}

\def\eqref#1{equation~\ref{#1}}

\def\1{\bm{1}}

\DeclareMathAlphabet{\mathsfit}{\encodingdefault}{\sfdefault}{m}{sl}
\SetMathAlphabet{\mathsfit}{bold}{\encodingdefault}{\sfdefault}{bx}{n}

\makeatletter
\DeclareRobustCommand{\cev}[1]{%
  {\mathpalette\do@cev{#1}}%
}
\newcommand{\do@cev}[2]{%
  \vbox{\offinterlineskip
    \sbox\z@{$\m@th#1 x$}%
    \ialign{##\cr
      \hidewidth\reflectbox{$\m@th#1\vec{}\mkern4mu$}\hidewidth\cr
      \noalign{\kern-\ht\z@}
      $\m@th#1#2$\cr
    }%
  }%
}
\makeatother

\usepackage{url}
\usepackage{footmisc}
\usepackage{caption}
\usepackage{subcaption}
\usepackage{xcolor}
\usepackage{tikz}
\usetikzlibrary{quantikz2} 
\usepackage{multirow} 
\usepackage{orcidlink} 
\usepackage{doi} 

\usepackage{color}
\usetikzlibrary{positioning, arrows, automata}
\usetikzlibrary{backgrounds, calc}
\usepackage{pgfplots}
\pgfplotsset{compat=1.8}
\usepackage{pgfplotstable}
\usepgfplotslibrary{groupplots}

\usepackage{microtype}
\usepackage{graphicx}
\usepackage{booktabs} 

\usepackage{hyperref}

\usepackage[accepted]{template}

\usepackage{amsmath}
\usepackage{amssymb}
\usepackage{mathtools}
\usepackage{amsthm}

\usetikzlibrary{positioning, arrows, automata, matrix}

\usepgfplotslibrary{groupplots}
\usepgfplotslibrary{colorbrewer}
\usetikzlibrary{external}
\usetikzlibrary{backgrounds, calc}
\usetikzlibrary{spy}
\usetikzlibrary{fit}
\usetikzlibrary{arrows.meta}
\usepackage{adjustbox}
\usepackage{graphicx}

\definecolor{lightgray204}{RGB}{204,204,204}
\definecolor{plot_background}{HTML}{EBF0F0}

\definecolor{C0}{HTML}{1f77b4}  
\definecolor{C1}{HTML}{ff7f0e}  
\definecolor{C2}{HTML}{2ca02c}  
\definecolor{C3}{HTML}{d62728}  
\definecolor{C4}{HTML}{9467bd}  
\definecolor{C5}{HTML}{8c564b}  
\definecolor{C6}{HTML}{e377c2}  
\definecolor{C9}{HTML}{17becf}  
\definecolor{C7}{HTML}{7f7f7f}  
\definecolor{C8}{HTML}{bcbd22}  
\definecolor{C10}{HTML}{CCCC00} 

\usepackage[capitalize,noabbrev]{cleveref}

\definecolor{nfvi}{RGB}{255,127,14} 

\theoremstyle{plain}
\newtheorem{theorem}{Theorem}[section]

\theoremstyle{definition}
\newtheorem{definition}[theorem]{Definition}

\theoremstyle{remark}

\usepackage[disable,textsize=tiny]{todonotes}

\icmltitlerunning{SQARL: A Size-Agnostic RL approach for Circuit Allocation in Distributed Quantum Architectures}

\begin{document}

\twocolumn[
\icmltitle{SQARL: A Size-Agnostic Reinforcement Learning approach for Circuit Allocation in Distributed Quantum Architectures}



\icmlsetsymbol{equal}{*}

\begin{icmlauthorlist}
\icmlauthor{V\'ictor Carballo}{1}{victorc@cs.upc.edu}{\orcidlink{0009-0009-4121-8982}}\hspace{15pt}
\icmlauthor{J\'ulia L\'opez-Closa}{2}{julia.lopez@bsc.es}{\orcidlink{0009-0000-7759-2591}}\hspace{15pt}
\icmlauthor{Mario Martin}{1}{mmartin@cs.upc.edu}{\orcidlink{0000-0002-4125-6630}}
\end{icmlauthorlist}

\icmlaffiliation{1}{Computer Science Department, Universitat Polit\`ecnica de Catalunya - BarcelonaTech (UPC)}
\icmlaffiliation{2}{High Performance Artificial Intelligence group, Barcelona Supercomputing Center}

\icmlkeywords{Reinforcement Learning, Combinatorial Optimization, Distributed Quantum Computing, Qubit Allocation, Transformer Architecture, Modular Quantum Architectures}

\vskip 0.3in
]

\printAffiliationsAndNotice{}  
\begin{abstract}
The scaling of quantum processors is currently limited by technical challenges such as decoherence and cross-talk.
As the number of qubits grows, interference increases the computational noise.
Distributed quantum computing addresses these limitations by interconnecting smaller, easier-to-handle quantum processors (cores), but it introduces the challenge of minimizing slow, error-prone inter-core communication.
The task of distributing quantum circuits across cores while minimizing communication costs is known as the~\emph{Qubit Allocation} problem.
This work focuses on developing a deep learning approach to this problem, emphasizing flexibility to quantum hardware topology and improving state-of-the-art performance.

Heuristic and non-learning algorithms, such as the Hungarian Qubit Allocation (HQA), currently represent the state of the art.
Reinforcement Learning (RL) approaches leverage learned allocation policies but often lack flexibility, requiring retraining when hardware configurations change, and they fall short of the solution quality achieved by non-learning methods.
However, learning mechanisms could outperform human-crafted heuristics.

To overcome these limitations, this work proposes a flexible, transformer-based architecture that can handle arbitrary numbers of qubits and cores without retraining.
Results show that the trained policy consistently outperforms the previous RL state of the art and narrows the gap between RL and HQA for the most common circuits. It achieves a 33\% reduction in allocation cost relative to the HQA for the Cuccaro Adder and 25\% on average for random circuits.
These findings show that learning-based approaches can effectively match the performance of hand-crafted heuristics, a crucial step towards their application in real-world scenarios.
\end{abstract}

\section{Introduction}
\label{sec:introduction}

Quantum computing has the potential to revolutionize numerous fields, including chemistry \cite{orobator2025applications}, cryptography \cite{mavroeidis2018impact}, computer science \cite{shor1999polynomial}, and many others. This promise, however, is constrained partly by the number of qubits a device can effectively control, which, even in the most advanced quantum computers available today, is on the order of a thousand \cite{gambetta2023, atomcomputing_ac1000}. Reaching the number of qubits required by real-world applications, on the order of millions \cite{preskill2018nisq}, is challenging due to decoherence, a phenomenon characterized by random fluctuations in qubit states and cross-talk between them. 

\begin{figure}[ht!]
    \centering
    \includegraphics[width=\linewidth]{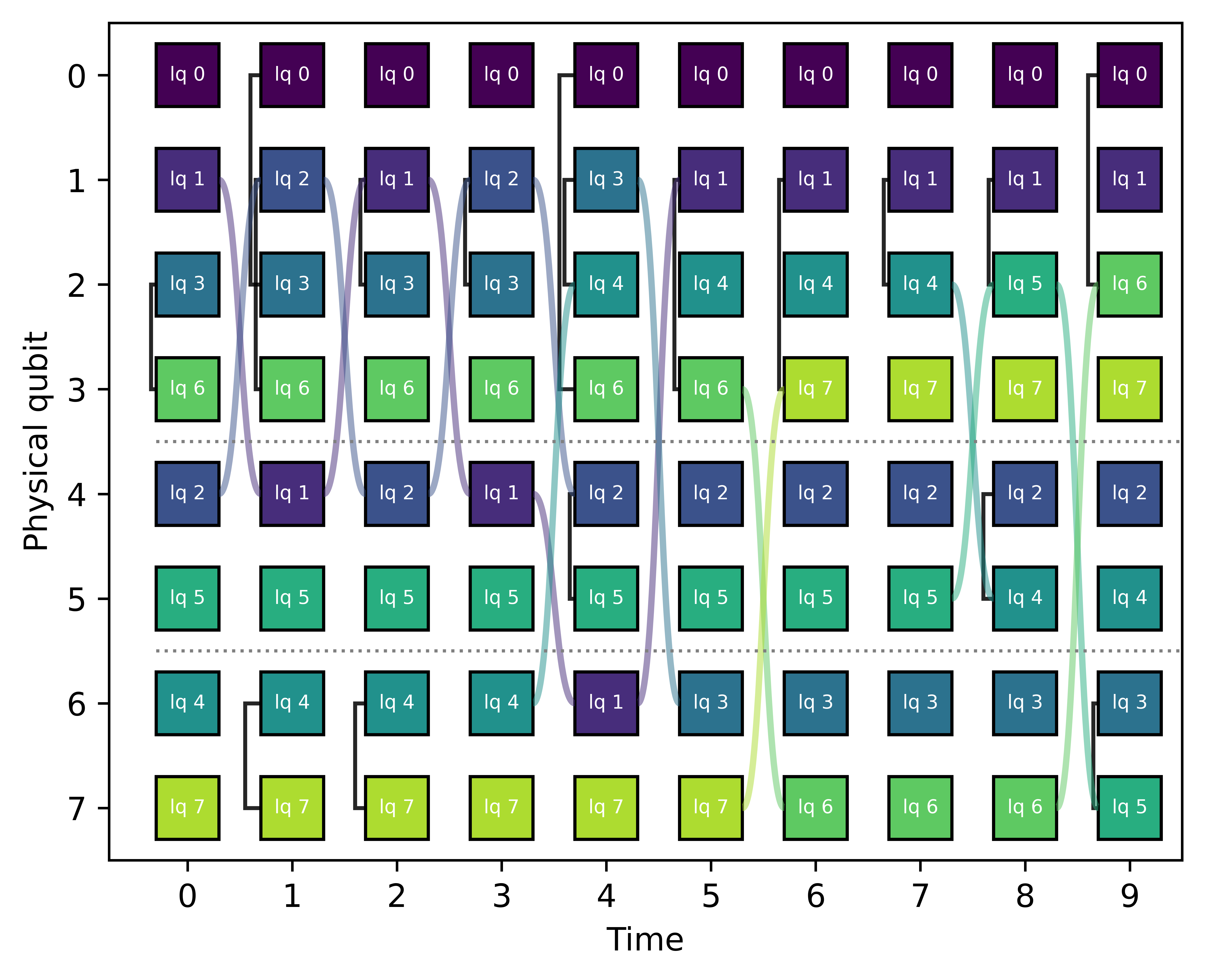}
    \caption{Illustrative example of circuit allocation. The $x$ axis represents the time steps required to execute the circuit. The $y$ axis represents each one of the actual qubits in the quantum hardware (physical qubits). The different processors in the distributed system are divided by dotted lines. The qubits in the original circuit (logical qubits) are labeled and color-coded; gates are drawn as black, vertical lines joining qubits.}
    \label{fig:alloc_ex}
\end{figure}

In traditional single-chip processors, scaling qubit counts is constrained by the growing complexity of control circuitry and wiring required to maintain low error rates \cite{mohseni2025buildquantumsupercomputerscaling}. As a response, multi-core architectures that aim to reduce cross-talk and interference while preserving the benefits of quantum computing have recently gained popularity \cite{vandersypen2017multicore, jnane2022multicore}. These devices consist of multiple interconnected quantum processors, also known as~\emph{cores}. The program to be executed is now distributed across smaller, more manageable quantum devices. Yet, multi-core architectures introduce their own challenges; in particular, reduced fidelity in non-local communications and additional time overhead when applying gates across cores \cite{rodrigo2021modellingSQ}. It is therefore imperative to distribute the quantum programs so as to minimize the number of expensive inter-core communications.
Fig.~\ref{fig:alloc_ex} shows an example of circuit allocation.

Quantum programs, also known as \textit{circuits}, are a high-level representation that specifies a sequence of operations on the qubits. Given a quantum circuit, qubit allocation is the mapping of the circuit's qubits (\emph{logical qubits}) to the actual qubits in the device's cores (\emph{physical qubits}), accounting for the device's hardware topology. Finding a mapping that minimizes communication is known to be NP-complete for single-chip processors \cite{siraichi2018qa, Botea2018OnTC}; similarly, minimizing inter-core operations in multi-core architectures leads to a combinatorial optimization problem. Hence, the existing literature often reformulates it as other well-known combinatorial problems, such as graph partitioning~\cite{baker2020time}, quadratic minimization~\cite{bandic2023mapping}, or resource allocation~\cite{escofet2023hungarian}, and then solves it using classical optimization algorithms.

Inspired by recent breakthroughs in deep learning, more recent works explore applying reinforcement learning (RL) to the problem \cite{pastor2024rl, russo2024attention}. RL methods train a policy to output a probability distribution over the assignment of qubits to cores. Despite the success of RL in fields like robotics control or table games, the solution quality of RL methods lags behind that of classical solvers such as the Hungarian algorithm. Furthermore, although not inherent to RL, existing approaches require retraining from scratch whenever the number of qubits or cores changes.
This is a crucial limitation to their applicability in real-world scenarios. 

In this paper, we present SQARL (Scalable Qubit Allocation via Reinforcement Learning), an RL-based approach that greatly narrows the gap in solution cost relative to classical optimization methods. In contrast to previous approaches, SQARL is unconstrained by circuit size and hardware configuration, thanks to its qubit- and core-agnostic architecture, enabling generalization across arbitrary hardware topologies and qubit counts without requiring retraining. One trains the model once, and it can be used for circuits of any size, which we define as flexibility. At the same time, SQARL achieves state-of-the-art performance on many quantum circuits and near–state-of-the-art performance on most of the remaining ones. To demonstrate these claims, we benchmark SQARL against the current SOTA classical algorithm, the Hungarian Qubit Allocation, proposed in~\cite{escofet2023hungarian}, and the best-performing existing RL-based approach, Russo's algorithm from~\cite{russo2024attention}.
Notably, while the algorithm is trained exclusively on small circuits (less than 20 qubits and 16 time slices), later benchmarks on much larger circuits (100 qubits and hundreds of time slices) show consistent performance on unseen problem sizes.
Overall, this work makes the following contributions:
\begin{enumerate}
    \item We compare the allocation cost of non-learning and state-of-the-art RL methods.
    \item We propose a novel, problem-size-agnostic policy architecture that can adapt to varying problem sizes without retraining.
    \item We propose a novel allocation procedure (sequential and parallel, as described in Sec.~\ref{subsec:alloc_proc}) that improves the algorithm's performance.
    \item We leverage a combination of REINFORCE and GRPO as training algorithms so as to enhance stability during training.
\end{enumerate}

The work is structured as follows.
In Sec.~\ref{sec:prel} we provide an introduction to quantum computing, the problem of qubit allocation, and RL with a special focus on combinatorial optimization.
Then, in Sec.~\ref{sec:related} we benchmark the state of the art in non-learning and RL algorithms, and discuss the performance gap and what needs to be improved in RL methods applied to this problem for them to be competitive.
Next, in Sec.~\ref{sec:methodology} we present our approach to quantum allocation via RL, SQARL.
Sec.~\ref{sec:results} covers the training procedure, hyperparameter selection, and benchmarks with both methods discussed in Sec.~\ref{sec:related}.
In Sec.~\ref{sec:discussion}, we discuss the results from the previous section.
Finally, in Sec.~\ref{sec:conclusion} we present our conclusions and future work.

The complete implementation is public and can be found in the project repository \url{https://github.com/Vicara12/SQARL}, along with the benchmarking circuits in JSON format \cite{Vicara12_MAI_TFM_2023}.

\section{Preliminaries}
\label{sec:prel}

RL approaches to quantum allocation require concepts from quantum computing and artificial intelligence. In this section, we provide the necessary background to understand the rest of the work.

\subsection{Quantum Computing}
\label{sec:intro_qc}

Quantum programs specify how quantum information is initialized, transformed, and measured in order to implement an algorithm. In the circuit model, this corresponds to a finite sequence of operations applied to the qubits prior to measurement.

Analogous to logic gates in conventional digital circuits, quantum gates are operators represented as unitary matrices that act on qubits by transforming their states. A quantum circuit is executed by applying a sequence of quantum gates to a set of qubits over time.

Quantum circuits are usually designed in an architecture-agnostic manner. As a result, they cannot be executed directly on a specific quantum device without accounting for hardware constraints.  
For instance, the set of possible quantum gates is infinite; yet, due to the challenges of gate calibration and control over many qubits, practical quantum hardware supports only a finite subset. From this \textit{universal} gate set, any quantum circuit can be approximated with arbitrary accuracy. Consequently, multi-qubit operations are compiled into sequences of one- and two-qubit gates supported natively by the target hardware. 

Another constraint is limited qubit connectivity: two-qubit gates can only be applied to physically connected qubits. The arrangement of qubits, also known as the qubit topology, can limit the execution of a circuit. In practice, quantum computers rely on inserting SWAP operations (i.e., information exchanges between two physical qubits) to relocate quantum states and enable the required interactions. These SWAPs, typically a sequence of three two-qubit CNOT gates, can only be applied to adjacent qubits in a chip; hence, such operations are inherently intra-core. In contrast, inter-core communications use the quantum teleportation mechanism, which has 5x to 100x longer latencies and incurs error rates 10x to 100x worse than two-qubit gates \cite{baker2020time, rodrigo2021scaling}. Therefore, it is imperative to reduce the number of expensive non-local communications when mapping a circuit to the hardware. 

This mapping of a quantum program to a specific gate set and connectivity is handled by a quantum compiler. More generally, the quantum compiler is responsible for all processes required to prepare a circuit for execution, including circuit synthesis and optimization, transpilation, qubit routing, and qubit allocation \cite{bandic2023mapping}. Efforts to minimize the effects of decoherence generally focus on one or more of these processes.
\begin{figure}[h!]
    \centering
    \includegraphics[width=0.99\linewidth]{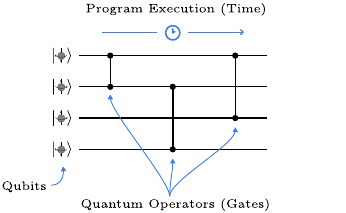}
    \caption{An annotated example of a quantum circuit.}
    \label{fig:circuit_ex}
\end{figure}

Quantum circuits are represented as a set of horizontal lines representing qubits, with symbols or vertical lines denoting the quantum gates applied to them. The circuit's execution proceeds in a left-to-right, chronological order. Figure \ref{fig:circuit_ex} shows an example of a quantum circuit. 

\begin{figure}[h!]
    \centering
    \includegraphics[width=0.6\linewidth]{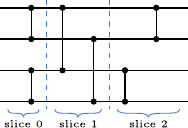}
    \caption{An example of circuit slicing.}
    \label{fig:slice_ex}
\end{figure}

Quantum gates that do not have qubits in common can be executed simultaneously. A time slice is a contiguous group of gates that do not share any qubits and can thus be executed at the same time. For instance, consider the circuit in Fig.~\ref{fig:slice_ex}, which can be divided in three different time slices.

\subsection{The Problem of Qubit Allocation}
\label{sec:problem}

Given a quantum circuit and a quantum hardware configuration, the problem of qubit allocation is to distribute the qubits across the different quantum cores for each circuit time slice in such a way that the total cost of the inter-core communications is minimized.

A time-sliced quantum circuit $G$ (refer to Sec.~\ref{subsec:circ_proc} for the slicing procedure) is a list of $T$ time slices $S_t,\ t\in \{0, ..., T-1\}$.
Each time slice $S_t$ is a list of gates, with no repeated qubits.
A gate is a two-element set $\{q_a,q_b\}$ that contains the qubits that participate in the gate.
The quantum hardware is defined by the number of cores $C$, the vector with the number of qubits each core can hold $\mathbf{c}$, and a matrix of inter-core communication costs $\mathbf{F}$.
The entry $\mathbf{F}(i,j)$ indicates the cost of moving a qubit from core $i$ to core $j$ in some defined metric (time, error probability, etc).
Although in real setups not all cores may be connected, we consider a fully connected cost matrix $\mathbf{F}$; if two cores are not directly connected, the corresponding entry in $\mathbf{F}$ contains the most efficient path through other cores.
The resulting allocation is a matrix of size $T \times Q$ over the set of cores $\{0, \dots, C-1\}$ that contains which core a qubit is assigned to in a specific time slice, that is $\mathbf{R} \in \{0, \dots, C-1\}^{T \times Q}$.
$\mathbf{R}(i,j) = k$ indicates that at time slice $i$, the qubit $j$ is at core $k$.

For the allocation to be valid, any pair of qubits participating in a gate within a given time slice $S_t$ must be assigned to the same core, so that a physical interaction between them is available.
We do not consider non-local communications during the execution of a time slice $S_t$; thus, qubit relocation is restricted to transitions between successive slices.
Another restriction is that any qubit allocation must respect each core's capacity.

Formally, we define the qubit allocation problem as follows: 

\begin{definition}[Qubit Allocation Problem]\label{def:qa}
    Given a quantum circuit $G$ over $Q$ qubits, consisting of $T$ time slices, and a hardware $\{\mathbf{c}, \mathbf{F}\}$ over $C$ cores, find a logical qubit to core per time slice assignment, that is, a matrix over the set of cores $\mathbf{R} \in \{0, \dots, C-1\}^{T \times Q}$ that minimizes the total inter-core communication cost
    \begin{equation}
        \sum_{0 < i < T} \sum_{0 \le k < Q} \mathbf{F}(\mathbf{R}(i-1,k), \mathbf{R}(i,k))
    \end{equation}
    subject to the following constraints:
    \begin{enumerate}
        \item Qubits in the same gate are in the same core:
        \begin{equation}
            \forall \{q_a, q_b\} \in S_t:\ \mathbf{R}(t, a) = \mathbf{R}(t, b).
        \end{equation}
        \item Core capacities are respected:
        \begin{equation}
            \begin{split}
                \forall S_t, j\in\{0, \dots, C-1\}:\ \ \ \ & \\
                |\{k \mid 0 \le k < Q, \mathbf{R}(t,k) & = j\}| \le \mathbf{c}_j.
            \end{split}
        \end{equation}
    \end{enumerate}
\end{definition}

\subsection{Reinforcement Learning}

In recent years, many works have explored Reinforcement Learning for combinatorial problems formulated as sequential decision processes \cite{bello2017neuralcomboptrl,kool2019attentionlearnsolverouting,rl4co}. Previous studies have also examined supervised (or imitation) learning for such problems \cite{drakulic2023bqncosupervised, luo2024ncosupervised}. However, obtaining labeled optimal solutions is infeasible in practice, since exact methods do not scale to large instances. On the other hand, heuristic methods designed to alleviate computational complexity often yield lower-quality solutions than exact methods; a model trained with imitation learning would then learn suboptimal behaviour. Instead, RL learns heuristics directly from the solution cost; that is, the neural network parameters are updated based on the outcome of each trial. This enables exploration of the state space and yields models that generate higher-quality solutions.

The construction process for a given instance $\mathbf{x}$ can be formulated as a Markov Decision Process (MDP) described by a tuple $\left(\mathcal{S}, \mathcal{A}, \mathcal{T}, \mathcal{R}, \gamma\right)$. The state $s_t\in \mathcal{S}$ represents the partial solution at time step $t$, and the action space $\mathcal{A}$ consists of all possible construction decisions. The transition function $\mathcal{T}$ deterministically updates the state based on the chosen action $a_t$, while $\mathcal{R}$ evaluates the constructed solution through rewards on each state transition $r_t = \mathcal{R}(a_t, s_t)$. Rewards are positive in most RL setups, whereas in cost optimization, they are typically set to the negative cost of actions. Thus, reward maximization is equivalent to cost minimization. The parameter $\gamma$ is a discount factor applied to rewards in future states; in optimization problems, it is usually set to 1.

In combinatorial optimization, RL policies are often parameterized using an encoder–decoder architecture, in which a problem instance $\mathbf{x}$ is first mapped to an embedding space $\mathbf{h}=f(\mathbf{x})$, and the decoder defines the conditional action distribution $\pi_\theta(a_t| s_t, \mathbf{h})$ used to construct the solution autoregressively~\cite{rl4co}.

The primary goal of this RL approach is to find the optimal parameters $\theta^*$ for the policy $\pi$ that maximizes the expected cumulative reward across all sampled problem instances. This objective is mathematically defined as:
\begin{equation}
\theta^*=\arg\max_\theta \mathbb{E}_{\mathbf{x}\sim P(\mathbf{x})}\Bigg[ \mathbb{E}_{\pi(a|\mathbf{x})}\Bigg[ \sum^{T-1}_{t=0}\gamma^t\mathcal{R}(s_t, a_t)\Bigg] \Bigg]
\end{equation}

From Definition \ref{def:qa}, the qubit allocation problem admits $C^{TQ}$ possible assignments, which for a 100-qubit circuit with 10 slices on a 10-core architecture is about $10^{1000}$. Many of these solutions violate problem constraints and are therefore infeasible. Hence, RL autoregressive methods often enforce feasibility through action masking \cite{atomcomputing_ac1000}. The state-dependent action set $\mathcal{A}(s_t)\subseteqq\mathcal{A}$ is formed by all feasible actions $a_t$ at state $s_t$. 

\section{Related Work}
\label{sec:related}

Given that qubit allocation is a fundamental issue in distributed quantum systems, numerous works have proposed solutions with varying allocation costs.
First approaches focused on applying well-known discrete optimization techniques to this new problem.
Among these, we can find graph partitioning methods such as FGP-rOEE~\cite{baker2020time}, quadratic optimization techniques such as QUBO~\cite{bandic2023mapping}, and resource allocation approaches such as HQA~\cite{escofet2023hungarian}.
Benchmarks of these non-learning algorithms~\cite{escofet2025revisiting} reveal that HQA consistently outperforms the other methods in terms of allocation cost, and is therefore the state-of-the-art.

More recent works explore the application of learned allocation heuristics via RL.
The state of the art in this domain is Russo's algorithm from~\cite{russo2024attention}.
It leverages graph neural networks for circuit codification, an attention mechanism for core assignment, and learning via practicing allocations.
However, some components of the architecture make it non-flexible with respect to hardware topology: before training, one selects a number of qubits and core configurations, and any change to either requires retraining the entire model from scratch.
This is an important limitation when applying allocation algorithms to real quantum systems.
\begin{figure}[h!]
    \centering
    \includegraphics[width=0.9\linewidth]{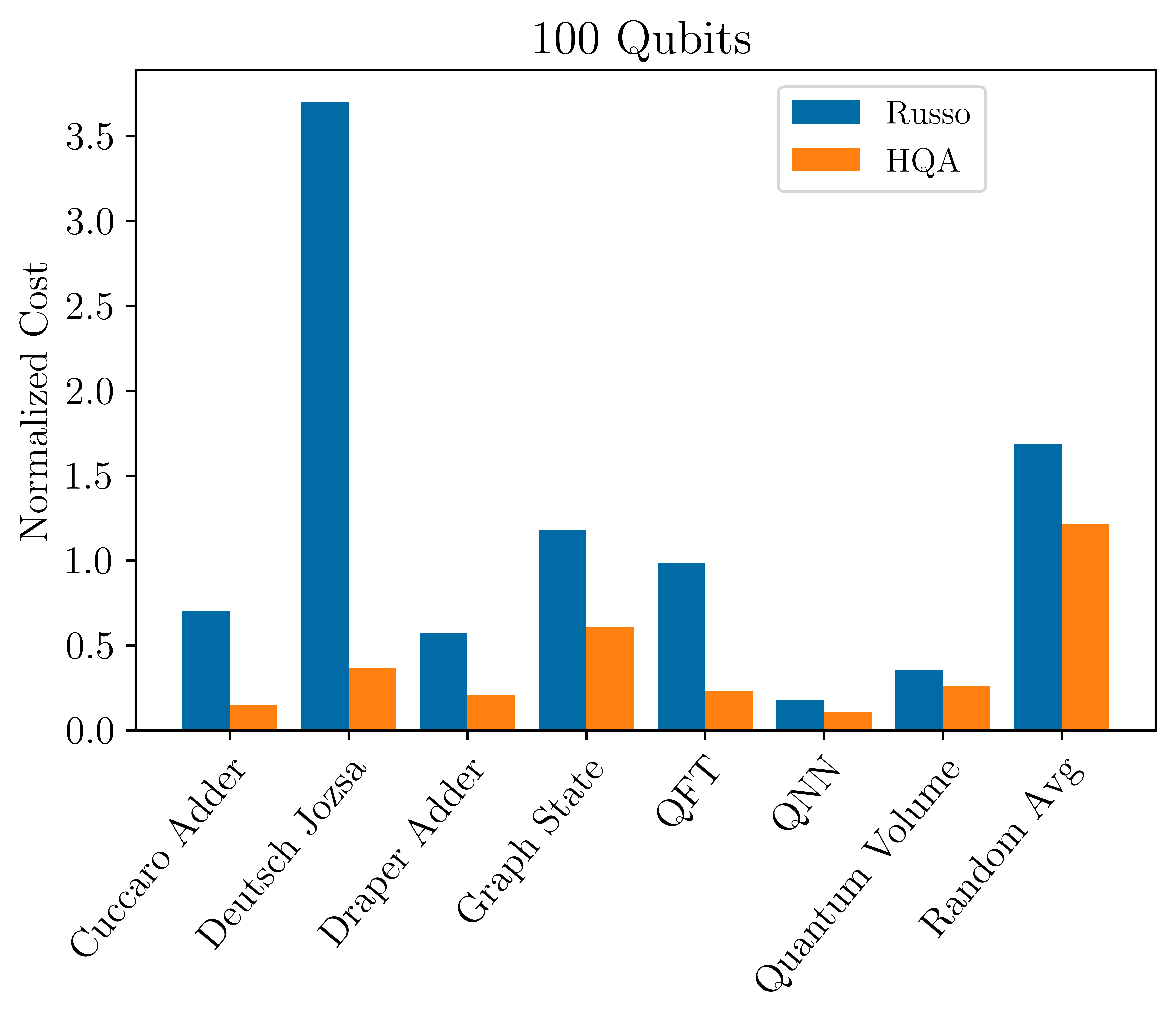}
    \caption{Allocation cost benchmark between the non-learning and RL state-of-the-art methods. Hardware consists of 10 quantum cores, each with 10 qubits. The intercommunication cost between any pair of cores is one unit.}
    \label{fig:sota_cost}
\end{figure}

No direct performance comparison in the literature exists between the non-learning and the RL state-of-the-art.
We present such a benchmark in Fig.~\ref{fig:sota_cost}, where both algorithms optimize a series of well-known quantum circuits, as well as the average cost over 64 random circuits, generated as described in Sec.~\ref{sec:training}.
The cost is reported as the number of inter-core communications per gate (excluding gates in the first time slice, as those would never require qubit relocation), to normalize values and facilitate easier comparison across circuits on the same scale.
There is a consistently large gap in the allocation costs between the two methods across all circuits, further impeding the use of RL techniques in real-world scenarios.

Despite the performance gap, we still consider RL a promising approach to qubit allocation.
The learning paradigm enables exploration-based improvement: the model can achieve better results by practicing different allocations.
In addition, RL has achieved state-of-the-art results in other domains, particularly in table games and robotics.
\section{Methodology}
\label{sec:methodology}

In this work, we adopt an RL approach.
By addressing its greatest disadvantages, the large cost gap relative to state-of-the-art non-learning methods and the size-flexibility constraints, RL algorithms could have practical use in real-world applications.

This section introduces SQARL (\emph{Scalable Qubit Allocation via Reinforcement Learning}), a novel algorithm for qubit allocation in distributed architectures.
First, in Sec.~\ref{subsec:circ_proc} we discuss the division of the circuit into time slices and the encoding of its information.
Then, in Sec.~\ref{subsec:pol_model}, we cover the policy model itself, its inputs, and the transformer-based architecture.
Following this, in Sec.~\ref{subsec:alloc_proc}, we present the two strategies we propose for circuit allocation --- sequential and parallel allocation.
Finally, in Sec.~\ref{sec:training}, we conclude with a description of the random circuit sampling method used for data generation during training, exploration techniques, and the training procedure.

\subsection{Circuit Encoding}
\label{subsec:circ_proc}

The unsliced circuit $G^\prime$ is received as an ordered list of two-qubit gates $\{q_a,q_b\}$, with $q_a \ne q_b$.
However, the procedure expects a time-sliced circuit $G$, formed by a list of $T$ time slices $S_t$, $t \in \{0,\dots,T-1\}$.
Each time slice is a list of two-qubit gates $\{q_a,q_b\}$, without qubit repetitions within each slice.
Thus, we need to group the consecutive gates from $G^\prime$ that can be executed simultaneously.

We use a simple, fast slicing method, shown in Alg.~\ref{alg:circuit_slicing}.
Gates are aggregated into a slice until one of the qubits interacts for a second time.
The gate containing that qubit forms a new time slice.
The set $\mathcal{U}$ keeps track of which qubits have already been used in the current time slice.

\begin{algorithm}[h!]
\caption{Circuit Slicing}
\small
\label{alg:circuit_slicing}
    \begin{algorithmic}
    \INPUT An unsliced circuit $G^\prime$.
    \OUTPUT Sliced circuit $G$.
    \STATE $G \gets []$\;
    \STATE $S_t \gets []$\;
    \STATE $\mathcal{U} \gets \emptyset$\;
    \FOR{Gate $\{q_a,q_b\}$ in $G^\prime$} 
        \IF{$q_a \in \mathcal{U} \lor  q_b \in \mathcal{U}$} 
            \STATE $G$\texttt{.append(}$S_t$\texttt{)}\;
            \STATE $S_t \gets []$\;
            \STATE $\mathcal{U} \gets \emptyset$\;
        \ENDIF
        \STATE $S_t$\texttt{.append(}$\{q_a,q_b\}$\texttt{)}\;
        \STATE $\mathcal{U} \gets \mathcal{U} \cup \{q_a,q_b\}$\;
    \ENDFOR
    \STATE $G$\texttt{.append(}$S_t$\texttt{)}\;
    \STATE \textbf{return} $G$\;
    \end{algorithmic}
\end{algorithm}

Some components of the algorithm do not use the list of gates and time slices directly; instead, they are encoded as a set of circuit features.
These aim to capture the same information in a form digestible to the model.
The first circuit feature is the~\emph{circuit embedding}, inspired by the lookahead weights from FGP-rOEE, proposed in~\cite{baker2020time}.
It contains, for each time slice, all qubit interactions from that slice until the end of the circuit.
A time slice $S_t$ can be encoded as an adjacency matrix $\mathbf{A}_t \in \{0,1\}^{Q \times Q}$, where $Q$ is the number of qubits in the circuit, as
\begin{equation}
    \mathbf{A}_t(i,j) = \begin{cases}
        1 & \text{if $\{q_i,q_j\} \in S_t$}\\
        0 & \text{otherwise}
    \end{cases}.
\end{equation}
The circuit embedding for slice $t$, $\mathbf{E}_t$, is a $\mathbb{R}^{Q \times Q}$ matrix defined as
\begin{equation} \label{eq:circ_emb}
    \mathbf{E}_t(i,j) = \sum_{t \leq k < T} \frac{1}{2^{k-t+1}} \mathbf{A}_k(i,j).
\end{equation}
These embeddings capture all the information in the circuit (the original set of gates can be reconstructed up to floating-point precision of $\mathbf{E}_t$).
However, the exponentially decaying sum of adjacency matrices can hinder the model's ability to capture relationships in distant qubit interactions.
This motivates a second circuit feature, the~\emph{next interaction} matrices.
These consist of a matrix $\mathbb{R}^{Q \times Q}$ for each circuit slice, defined as
\begin{equation}
    \mathbf{N}_t(i,j) = \underset{t \le k < T}{\max} \frac{T - k}{T - t} \mathbf{A}_k(i,j).
\end{equation}
$\mathbf{N}_t(i,j)$ contains a linearly decaying value that indicates how close the next interaction of qubits $q_i$ and $q_j$ is --- $1$ if it happens in this slice and $0$ if they no longer interact.

\subsection{The Policy Model}
\label{subsec:pol_model}

The policy model, $\pi_\mathbf{\theta}$, is the neural network component of the algorithm.
It is used during the allocation procedure to determine to which core a qubit is assigned.
The allocation procedure is discussed in-depth in Sec.~\ref{subsec:alloc_proc}, but in essence, qubits are allocated in each time slice using the policy.
If two qubits participate in a gate, they are allocated together; otherwise, they are allocated individually.
The policy provides a probability distribution over the set of cores at each allocation step, and qubits are allocated to cores according to this distribution.
Once the policy is trained, allocating qubits to high-probability cores yields lower-cost solutions.

The policy model must receive all necessary information to determine this probability distribution as effectively as possible.
This information is condensed in a vector of features for each of the $C$ hardware cores and $Q$ qubit combinations, producing a tensor of shape $[B,C,Q,h]$, where $B$ is the batch size and $h$ is the number of features, in our case ten.

Let $q_a$ and $q_b$ be the two qubits being allocated in a time step.
The first feature is a one-hot encoding indicating which qubits are being allocated.
The tensor element at position $[b,c_i,q_j,0]$ equals 1 if $q_j = q_a$ or $q_j = q_b$, zero otherwise.

The second feature indicates whether qubit $q_j$ was allocated to core $c_i$ in the previous time slice: 1 if so, 0 if not.
Similarly, the third feature provides information on current allocations.
Considering all qubits allocated thus far in the current slice, position $[b,c_i,q_j,2]$ is 1 if qubit $q_j$ is allocated to core $c_i$ in this time slice and 0 if not or it has not been allocated yet.

The fourth feature encodes the number of free spots in the $i$th core at this specific allocation step, $\mathbf{c}_i$.
It is encoded as $1/(\mathbf{c}_i + 1)$.
The fifth feature captures the total cost of moving $q_a$ and $q_b$, the qubits allocated, to core $c_i$, or just $q_a$ for single qubit allocations.
Analogously to core capacities, it is encoded as $1/(f + 1)$, where $f =\mathbf{F}(c^\prime_a, c_i) + \mathbf{F}(c^\prime_b, c_i)$, and $c^\prime_a$ ($c^\prime_b$) the core to which $q_a$ ($q_b$) was allocated in the previous time slice, or $f =\mathbf{F}(c^\prime_a, c_i)$ for single qubit allocations.
Both feature encoding styles as $1/(x + 1)$ enable the mapping of the range $[0,+\infty)$ to the bounded space $[1,0)$.

The sixth feature is qubit-to-core attraction, a concept first introduced in~\cite{escofet2023hungarian}.
The attraction between qubit $q_j$ and core $c_k$ at the $t$th time slice is defined as
\begin{equation} \label{eq:attr}
    \text{\texttt{attr}}_t(q_j,c_k) = \sum_{0 \le i < Q} \mathbf{J}_t(q_i, c_k) \cdot \mathbf{E}_t(q_j,q_i),
\end{equation}
where $\mathbf{J}_t(q_i, c_k) = 1$ if qubit $q_i$ is in core $c_k$ at time slice $t$, 0 otherwise,
and $\mathbf{E}_t$ is the circuit embedding as defined in Eq.~\ref{eq:circ_emb}.
The seventh and eighth features are the circuit embedding items, as defined in Eq.~\ref{eq:circ_emb}, at positions $\mathbf{E}_t(q_j,q_a)$ and $\mathbf{E}_t(q_j,q_b)$, respectively.
If the allocation is for a single qubit, then $\mathbf{E}_t(q_j,q_b)$ is replaced with 0.
Finally, the ninth and tenth features are the next interaction elements at positions $\mathbf{N}_t(q_j,q_a)$ and $\mathbf{N}_t(q_j,q_b)$.
Analogously, the tenth feature is replaced with 0 in single-qubit allocations.

\begin{figure*}[t!]
    \centering
    \includegraphics[width=\linewidth]{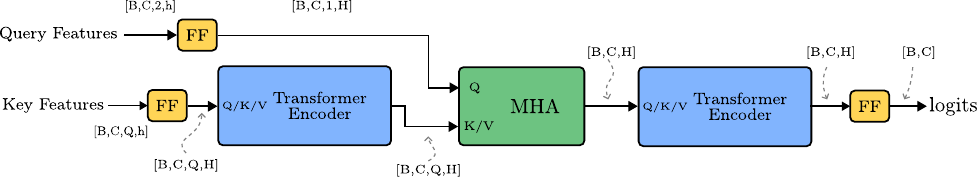}
    \caption{Policy model's architecture.}
    \label{fig:pred_model}
\end{figure*}

To summarize, when running the policy to allocate qubits $q_a$ and $q_b$, the feature tensor contains the following vector of elements at position $[b,c_i,q_j]$:
\begin{enumerate}
\itemsep0em 
    \item 1 if $q_j \in \{q_a,q_b\}$, 0 otherwise.
    \item 1 if $q_j$ was allocated to core $c_i$ in the previous time slice, 0 otherwise.
    \item 1 if $q_j$ is allocated to core $c_i$ in the current time slice, 0 otherwise.
    \item $1/(\mathbf{c}_i + 1)$, with $\mathbf{c}_i$ the number of free places in core $i$.
    \item $1/(f + 1)$, with $f$ the cost of moving $q_a$ and $q_b$ to core $i$.
    \item Average attraction of $q_a$ and $q_b$ to core $i$, as defined in Equation~\ref{eq:attr}.
    \item $\mathbf{E}_t(q_j,q_a)$.
    \item $\mathbf{E}_t(q_j,q_b)$ or 0 for single-qubit allocations.
    \item $\mathbf{N}_t(q_j,q_a)$.
    \item $\mathbf{N}_t(q_j,q_b)$ or 0 for single-qubit allocations.
\end{enumerate}
As noted previously, the features are analogous for single qubit allocations, considering the qubit $q_a$ alone.

The previous tensor contains features for all $Q$ qubits.
However, we can focus on the qubits being allocated in this step by selecting and extracting the relevant slices along the qubit dimension of the tensor, resulting in a new tensor of shape $[B,C,2,h]$, which contains feature vectors only for these qubits and all cores at this allocation step.
For single-qubit allocations, the tensor corresponding to that qubit is concatenated with zeros in the qubit dimension to get the desired shape.
We refer to the tensor over all qubits as the~\emph{key} tensor, and to the one over the qubits being allocated alone as the~\emph{query} tensor.
These are fed into the architecture shown in Fig.~\ref{fig:pred_model}.
Both the query and the key are projected from a feature space of size $h$ to another of size $H$, a hyperparameter of the model, using distinct linear projections.
The two vectors of features corresponding to the two qubits being allocated are also combined into a single one.
Then, the projected key features attend to one another through a transformer encoder block.
Next, the query and key features are combined using a~\emph{Multi Head Attention} (MHA) block.
In this step, the attention between the query and key tensors aggregates the information of all qubits in each core, producing a new tensor of features for the cores of shape $[B,C,H]$.
Lastly, the core features are allowed to attend to one another via another transformer encoder block, and are finally converted into a scalar via a linear projection.
The output of this model is a tensor of shape $[B,C]$, corresponding to the logits for the probability distribution over the set of cores for each element in the batch.

\subsection{The Allocation Procedure}
\label{subsec:alloc_proc}

Our algorithm uses an autoregressive approach to circuit allocation.
Slices are allocated sequentially, and within each slice we place one qubit at a time, or two if these form a gate.
Not all qubit-to-core assignations are valid solutions, as these must respect the hardware restrictions mentioned in Sec.~\ref{sec:problem}.
The probabilities of actions that would lead to illegal states need to be masked to prevent such solutions.
To minimize the number of situations requiring an intervention, we propose two conditions for the allocation order.
First, qubits that form a gate are allocated together.
This prevents them from being allocated to different cores.
Second, qubits that belong to gates are allocated first, and then unpaired qubits.
This prevents configurations in which no core has enough space for two qubits that must be allocated together.
Thus, the only illegal action that remains to be masked is assigning a qubit to a core withs no free spots left.

Slices are allocated in order; however, qubits do not have an intrinsic order in which they must be allocated.
Qubit allocation order could affect the algorithm's performance --- some qubits have stronger preferences for assignment than others.
We propose two allocation orders: a~\emph{sequential} order, in which a heuristic determines which qubits are allocated first; and a~\emph{parallel} order, in which the policy itself determines allocation priority.

\begin{figure}[htbp]
  \centering
  \begin{subfigure}[b]{0.2\textwidth}
    \centering
    \includegraphics[width=0.275\linewidth]{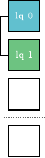}
    \caption{}
  \end{subfigure}
  \begin{subfigure}[b]{0.2\textwidth}
    \centering
    \resizebox{1.1\linewidth}{!}{%
    \begin{quantikz}
       \lstick{\text{lq 0}} & \ctrl{1}\slice{} & \ctrl{3} & \\
       \lstick{\text{lq 1}} & \phase{}         &          & \\
       \lstick{\text{lq 2}} &                  &          & \\
       \lstick{\text{lq 3}} &                  & \phase{}  & \qw
    \end{quantikz}
    }
    \caption{}
  \end{subfigure}

  \caption{A scenario where the free qubits’ order of allocation influences final cost. (a) Partial allocation
    diagram for hardware with two cores, with three and one qubit respectively, before assigning a core
    to logical qubit 2, first time slice only. (b) Circuit used in the example.}
  \label{fig:allocation_order_example}
\end{figure}

In sequential allocation, the qubits are pre-ordered according to a heuristic that prioritizes more active or earlier-interacting qubits.
Qubits that belong to some gate in the current time slice, allocated in pairs, are ordered according to their respective value in the circuit embedding tensor (i.e. $E_t(q_a,q_b)$ for qubits $\{q_a,q_b\}$ that form a gate in the $t$th time slice) in descending order.
Free qubits, those that do not participate in any gate in the current time slice, are sorted in descending order according to the maximum element in the embedding tensor row: for qubit $q_a$ and the $t$th time slice, we have $\max_{0\le i < Q} E_t(q_a, q_i)$.
Fig.~\ref{fig:allocation_order_example} shows an example in which this heuristic improves allocation cost.
Qubits 0 and 1 have been allocated; if qubit 2 were to come next, a naive policy could place it in the first core, ignoring that qubit 3 would greatly benefit from this spot, as it interacts with qubit 0 in the next time slice.
By allocating qubits that interact sooner first, these preferences are more easily taken into account: qubit 3 would be allocated before qubit 2.

\begin{figure}[htbp]
  \centering
  \begin{subfigure}[b]{0.25\linewidth}
    \includegraphics[width=\linewidth]{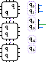}
    \caption{}
  \end{subfigure}\hfill
  \begin{subfigure}[b]{0.25\linewidth}
    \includegraphics[width=\linewidth]{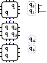}
    \caption{}
  \end{subfigure}\hfill
  \begin{subfigure}[b]{0.25\linewidth}
    \includegraphics[width=\linewidth]{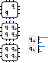}
    \caption{}
  \end{subfigure}\hfill

  \caption{Qubit allocation process.
  (a) Paired qubits are allocated together and before free qubits. The policy is applied to pairs $\{q_0, q_1\}$ and $\{q_2, q_3\}$ and, after sampling, the pair $\{q_2, q_3\}$ is assigned to the middle core.
  (b) After pair $\{q_2, q_3\}$ is taken out of the allocation pool, the policy is run again on the remaining pairs.
  (c) Once there are no more paired qubits left, the same process is repeated with the free qubits.
  }
  \label{fig:alloc}
\end{figure}

Parallel allocation aligns more closely with RL: replacing human-made heuristics with learned policies.
Thus, we decide the allocation order by using the policy itself.
All qubits to be allocated in a given slice are fed to the policy in batch.
The policy produces logits for all cores and qubits, which are then flattened into a single vector and fed to a softmax.
This produces a probability distribution over all qubit-to-core assignments; sampling from this probability distribution means allocating first the qubits with the highest core preferences.
When a qubit-core pair is selected, the given qubit is allocated to the core and taken out of the allocation pool.
The process is repeated until no qubits remain to be allocated.
Note, however, that paired qubits (those that belong to gates) are allocated first, and then free qubits.
The process is executed twice per time slice --- first exclusively for qubit pairs, then for free qubits.
Fig.~\ref{fig:alloc} shows an example of this process applied to the allocation of two pairs of qubits and two free qubits to a three-core hardware.

As will be shown in Sec.~\ref{sec:results}, sequential allocation does best in some circuits, whereas parallel does better in others.
Thus, we propose allocating each circuit using both methods and selecting the allocation that yields a lower overall cost.

\subsection{Training Procedure}
\label{sec:training}

The policy is trained by practicing allocations: a random circuit and hardware are generated, the circuit is optimized several times, and actions that lead to lower allocation costs are incentivized, whereas those that result in higher costs are penalized.
This process is repeated until the policy converges.

The first step is producing random circuits and hardware configurations.
The hardware-sampling procedure randomly selects a number of cores $C$ and a vector of core capacities $\mathbf{c}$ in a given range.
The model is trained with a uniform interconnection cost matrix $\mathbf{F}$, where moving a qubit between any two cores has a cost of one unit.
This is not a requirement, and new versions of the model could be trained on randomly sampled patterns of inter-core communication costs.
Analogously to hardware sampling, we sample from a uniform distribution over all possible circuits to achieve a circuit-agnostic policy.
On each iteration, a number of qubits $Q$ and time slices $T$ is selected randomly.
Then, gates $\{q_a,q_b\}$, $a,b \in \{0,\dots,Q-1\} \land a\ne b$, are sampled and fed to Alg.~\ref{alg:circuit_slicing} until the desired number of time slices $T$ is reached.

A crucial point in any RL algorithm is that of balancing exploration and exploitation.
This algorithm includes two exploration mechanisms.
The first is a vector of random positive values that is added to the probabilities produced by the policy.
Precisely, we have that
\begin{equation} \label{eq:noise}
    \mathbf{p}^\prime = \alpha \cdot \mathbf{x} + (1 - \alpha) \cdot \mathbf{p},
\end{equation}
where $\mathbf{p}$ is the original vector of probabilities, $\mathbf{x} \sim \mathcal{U}(0,1)^n$ a vector of random values sampled from a uniform distribution, and $\alpha$ a parameter that controls the noise-to-signal ratio.
This parameter decreases slightly with each iteration.
The second exploration mechanism is that of sampling distributions.
When executing SQARL at test time, the different distributions are not sampled; the highest-probability value is always selected, resulting in a greedy allocation.
This is to minimize allocation cost and perform deterministic allocations.
However, during training, it is beneficial to trade off some performance to obtain variety in the results.
The variety in actions and costs is used to learn better allocation strategies.,

Action sampling, unlike greedy selection, is also relevant for another component of the training algorithm --- advantage calculation.
Some implementations of the REINFORCE algorithm use a previous version of the algorithm as a baseline to measure advantage.
However, in combinatorial optimization problems, where small differences in allocations can yield large differences in value, this can lead to high variance and less stable training.
Thus, we propose an approach to advantage calculation borrowed from GRPO~\cite{shao2024deepseekmath}: a circuit is optimized several times, sampling actions from the policy's probability distributions, and the individual advantage of each allocation is obtained relative to the rest.
Specifically, the costs of all allocations are gathered into a vector $\mathbf{A}_c$, which is then normalized as
\begin{equation}
    \tilde{\mathbf{A}}_c = \frac{\mathbf{A}_c - \mu}{\sigma},
\end{equation}
where $\mu$ and $\sigma$ are the mean and standard deviation of $\mathbf{A}_c$.
This relative cost is then used in a training setup similar to the REINFORCE algorithm~\cite{williams1992simple}.

Another key aspect of RL is how to deal with illegal actions. In Sec.~\ref{subsec:alloc_proc}, we presented an allocation sequence that minimizes the amount of masking required.
However, experiments show that when masking is used during training, the model is unable to learn which actions are legal.
In the late stages of training, most of the probability mass is assigned to illegal actions.
Thus, it is important to teach the model which actions are possible, so that the probability of performing illegal actions is minimized and training does not diverge towards attempting illegal actions.
During training, the policy is allowed to perform illegal actions (allocating qubits to cores that lack empty slots, even if this leads to an invalid solution), but it is penalized.
Illegal actions are always masked outside training, despite these having very little probability mass once the policy is trained.

The training procedure is shown in Alg.~\ref{alg:rl_alg}. The parameter $\beta$ controls the impact of the illegal action loss.

\begin{algorithm}[h!]
\small
\caption{Reinforcement Learning Training}
\label{alg:rl_alg}
\begin{algorithmic}
\INPUT {Policy model $\pi_\theta$ with parameters $\theta$}
\WHILE{not converged} 
    \STATE $\{\mathbf{c},\mathbf{F}\} \gets 
    \text{\texttt{randomHardwareSampler}()}$\;
    \STATE $G \gets \text{\texttt{randomCircuitSampler}()}$\;
    \STATE $\{R_0, R_1, ..., R_{N-1}\} \gets \text{\texttt{optimizeNTimes}($G,\mathbf{c},\mathbf{F}, \theta$})$\;
    \STATE $\{\tilde{A}_{c_0}, ...\} \gets \text{\texttt{getNormalizedCostVector}($\{R_0,...\}$)}$\;
    \FOR{$n \in \{0,\dots,N-1\}$} 
        \FOR{allocation step $s_i$ and action $a_i$ in $R_n$} 
            \IF{\texttt{legal}($a_i,s_i$)} 
                \STATE loss $\gets \text{loss} + (1-\beta) A_n \log(\pi_\theta(a_i|s_i))$\;
            \ELSE
                \STATE loss $\gets \text{loss} + \beta \log(\pi_\theta(a_i|s_i))$\;
            \ENDIF
        \ENDFOR
    \ENDFOR
    \STATE $\theta \gets \text{\texttt{updateParameters}(loss,$\theta$)}$\;
\ENDWHILE
\end{algorithmic}
\end{algorithm}

\section{Experiments and Results}
\label{sec:results}

This section begins by reviewing the selected hyperparameters for the model and the training procedure.
Next, the trained policy is tested on a selection of circuits for sequential and parallel allocation.
Finally, we present a benchmark of the performance with the state of the art.
The code was executed in BSC's Marenostrum 5, with an Intel Xeon Platinum 8460Y+ CPU and a NVIDIA H100 GPU.

Several hyperparameters need to be fixed before training the architecture described in Sec.~\ref{subsec:pol_model}.
Specifically, we set the hidden embedding size $H$ to 64, the number of layers per transformer encoder block to 2, and the number of heads to 2, yielding a total of 119617 trainable parameters.
We trained the model for 28100 iterations on quantum hardware with 2 to 8 cores and circuits of up to 20 qubits and 4 to 16 time slices.
The size of the circuits and hardware used for training is relatively small; later, we will evaluate the policy on quantum circuits with 100 qubits and hundreds of time slices on hardware with 5 and 10 cores.
These samples fall outside the problem sizes the algorithm has seen during training, thereby testing its ability to scale to problem sizes beyond those encountered during training.
The group used in the GRPO advantage calculation is of size 32.
The invalid move penalty ($\beta$ in Alg.~\ref{alg:rl_alg}) is set to 0.3 for the first 16000 iterations and to 0.5 for the last 12100.
The initial noise injected into the policy's probabilities ($\alpha$ in Eq.~\ref{eq:noise}) is set to 0.2 with a decay factor of 0.999.
The noise ratio $\alpha$ was raised to 0.05 at iteration 16000.
The policy was trained with the sequential allocation policy, as it is more stable than parallel allocation for training.
The training lasted 124 total hours of computing.

\begin{figure}[h!]
    \centering
    \includegraphics[width=\linewidth]{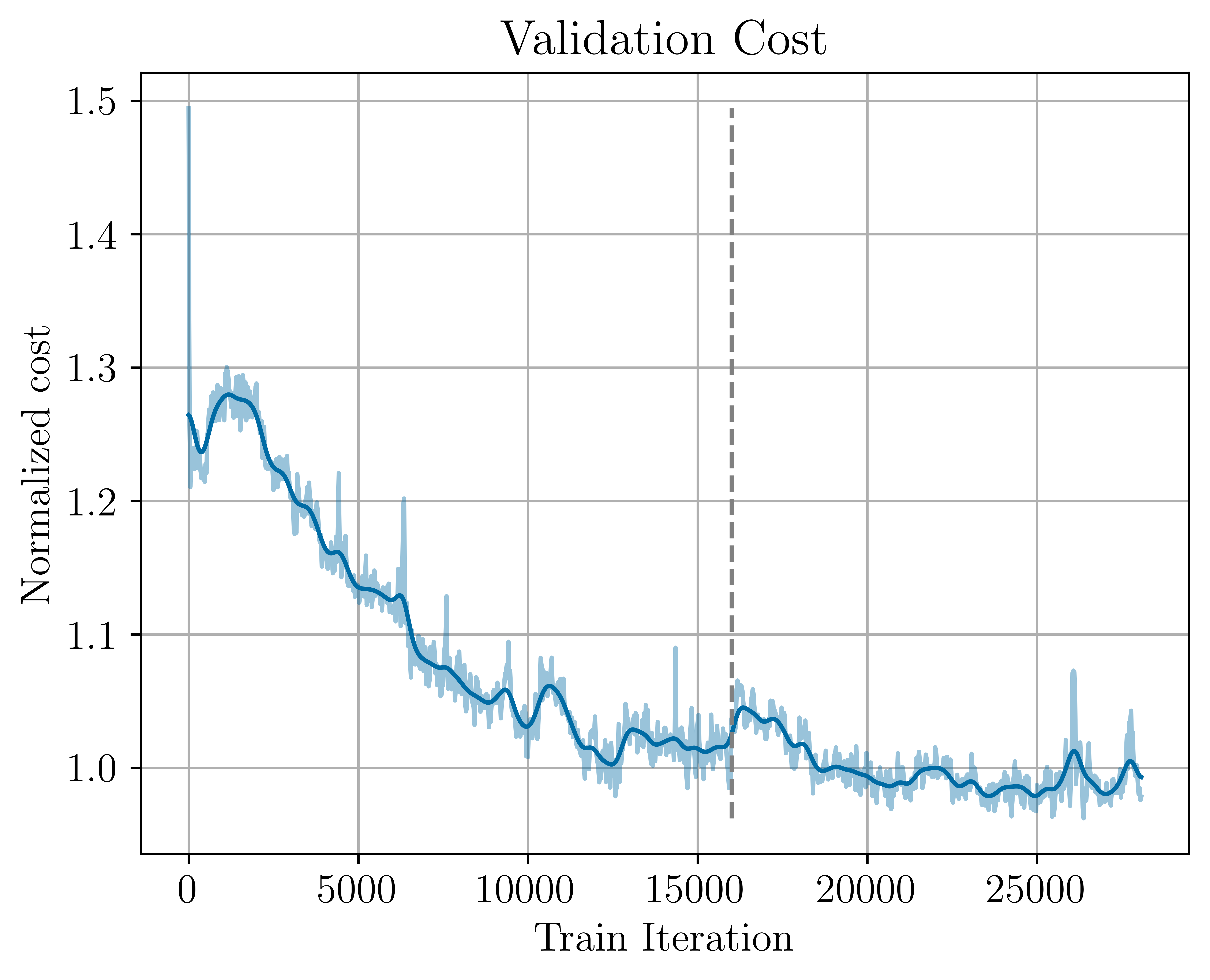}
    \caption{
    Normalized allocation cost on the validation set of circuits during training.
    }
    \label{fig:val_cost}
\end{figure}

Validation runs are executed every 25 iterations.
The average normalized cost during training is shown in Fig.~\ref{fig:val_cost}.
The gray dotted line indicates the point at which noise was added to the policy to incentivize further exploration.

\begin{figure}[h!]
    \centering
    \includegraphics[width=\linewidth]{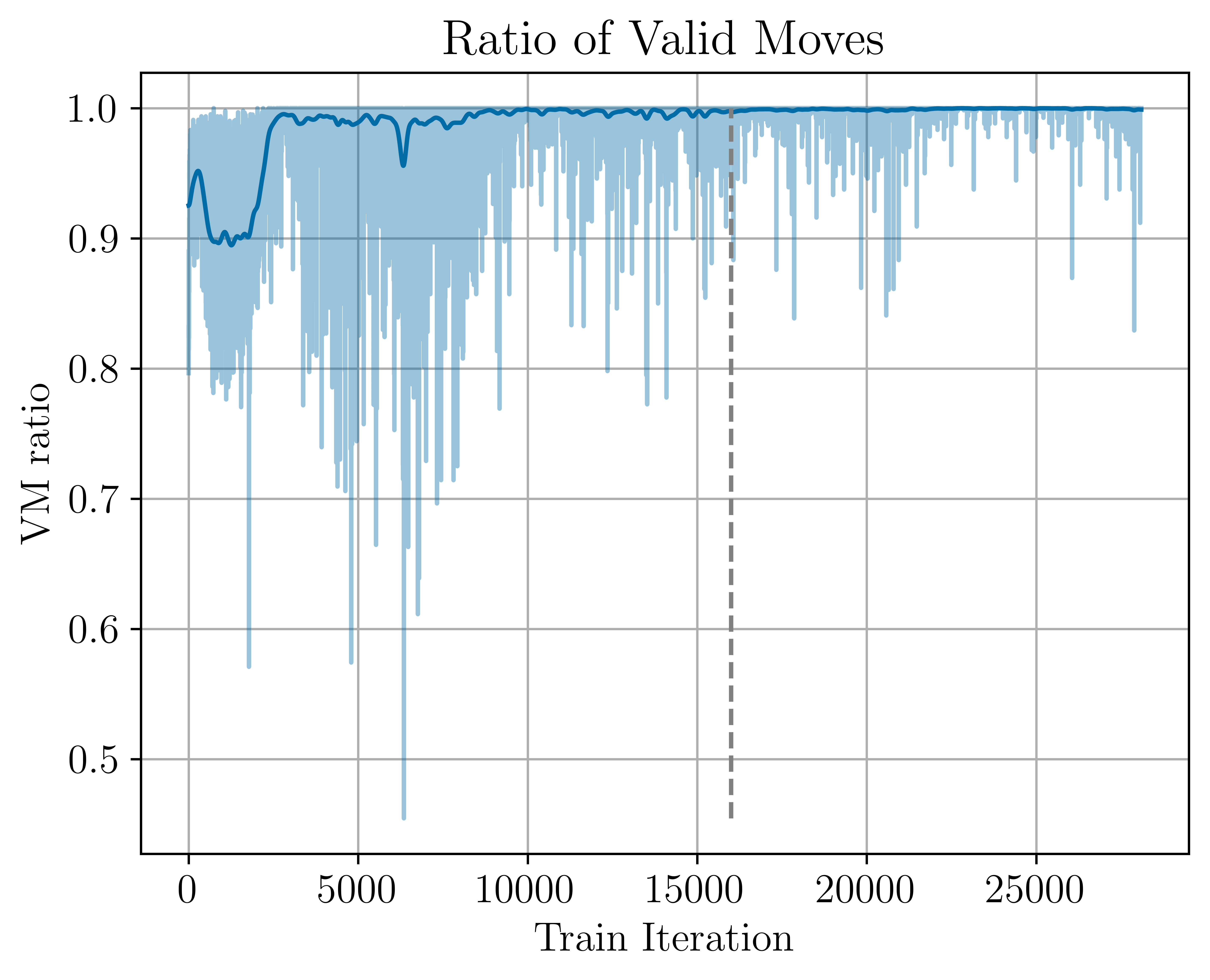}
    \caption{
    Ratio of valid moves selected by the policy during training.
    Data shown with a Gaussian filter ($\sigma = 10$) to smooth noise.
    }
    \label{fig:vm}
\end{figure}

In Sec.~\ref{sec:training}, we noted that during training, the policy is allowed to take illegal actions but is penalized for doing so.
If the invalid movement penalty is not set high enough, the policy could collapse into taking mostly illegal actions, as the performance gains from them outweigh the loss penalty.
Fig.~\ref{fig:vm} shows the ratio of valid moves as training progresses.
The policy rapidly learns which actions are illegal and takes mostly legal actions.
Furthermore, because of probabilistic action sampling during training, illegal actions may still be taken even when their probability is very small.
All illegal actions are masked when the model is not being trained.

For benchmarking, we have sampled 64 random circuits, generated as described in Sec.~\ref{sec:training}, with 50 time slices (whose costs are averaged under the category ``Random Avg’') and 7 relevant circuits in the field of quantum computing.
The Cuccaro Adder is a popular circuit for adding quantum registers.
It is highly efficient and requires few auxiliary qubits.
Deutsch-Jozsa is a quantum algorithm of great historical importance, as it was the first to show exponential quantum advantage over non-quantum algorithms.
It is used to tell whether a function is constant or balanced.
The Draper Adder is a circuit that performs addition in the phase basis using the quantum Fourier transform.
A Graph State is a highly entangled state represented by a graph, in which the vertices correspond to qubits and the edges represent entanglement.
It is the foundation of measurement-based quantum computing.
The Quantum Fourier Transform (QFT) is the quantum analogue of the discrete Fourier Transform and is used in some of the most famous quantum algorithms.
Quantum Neural Networks (QNN) are variational circuits with tunable parameters.
It is the primary candidate for near-term applications of quantum computers.
Quantum Volume is a benchmark protocol that runs random square circuits of increasing size on a quantum computer.
It is the industry standard for measuring hardware performance.

We first compare the sequential and parallel allocation modes in Tab.~\ref{tab:cost_my_versions} to determine which performs best.
Results are shown for two groups of circuits: one version with 50 qubits, optimized on a quantum computer with 5 cores, each with 10 qubits, and the other with 100 qubits, optimized on a quantum computer with 10 cores, each with 10 qubits.
No allocation mode consistently outperforms the other; we propose optimizing each circuit with both modes and selecting the solution with the lower cost in an ensemble-like manner.
Thus, the cost of SQARL is the minimum of the two, and the execution time is the sum of both.

\begin{table}[h]
\centering
\footnotesize
\begin{tabular}{|l|cc|cc|}
\hline
\multirow{2}{*}{\textbf{Circuit Name}} & \multicolumn{2}{c|}{\textbf{50 Qubits}} & \multicolumn{2}{c|}{\textbf{100 Qubits}} \\
& \textbf{Seq.} & \textbf{Par.} & \textbf{Seq.} & \textbf{Par.} \\\hline
Cuccaro Adder  & 52   & \textbf{50}   &   \textbf{74} &   76 \\
Deutsch Jozsa  & \textbf{16}   & \textbf{16}   & \textbf{36}  & 38  \\
Draper Adder  & \textbf{332}  & 348  &  \textbf{841} &  919 \\
Graph State    & \textbf{418}  & 476  & \textbf{1990} & 2079 \\
QFT            & \textbf{323}  & 424  & \textbf{1586} & 2506 \\
QNN            & \textbf{671}  & 848  & \textbf{3349} & 4182 \\
Quantum Volume & 1000 & \textbf{874}  & 4680  & \textbf{4241} \\
Random Avg     & 195  & \textbf{163}  &  348  &  \textbf{273} \\\hline
\end{tabular}
\caption{
  Comparison of the allocation cost for the two versions of the algorithm, sequential and parallel mode, in a selection of circuits with 50 and 100 qubits.
}
\label{tab:cost_my_versions}
\end{table}

Tab.~\ref{tab:perf_comp_sota} presents a cost-performance benchmark of our algorithm against the RL state of the art, Russo's algorithm, and the overall state of the art, HQA.

\begin{table}[h!]
\footnotesize
\begin{subtable}[b]{0.48\textwidth}
    \caption{}
    \small 
    \setlength{\tabcolsep}{2.2pt}
    \centering
    \begin{tabular}{|lccccc|}
        \hline
         & \multicolumn{3}{c}{\textbf{Inter-core Comms.}} & \multicolumn{2}{c|}{\textbf{Improvement}} \\
         & \textbf{Russo} & \textbf{HQA} & \textbf{SQARL} & \textbf{vs Russo} & \textbf{vs HQA} \\\hline
        Cuccaro Adder & 143 & 50 & 50 & \textbf{+65.03\%} & \textbf{+0.00\%} \\
        Deutsch Jozsa & 48 & 16 & 16 & \textbf{+66.67\%} & \textbf{+0.00\%} \\
        Draper Adder & 691 & 347 & 332 & \textbf{+51.95\%} & \textbf{+4.32\%} \\
        Graph State & 546 & 341 & 418 & \textbf{+23.44\%} & -22.58\% \\
        QFT & 1146 & 309 & 323 & \textbf{+71.82\%} & -4.53\% \\
        QNN & 573 & 546 & 671 & -17.10\% & -22.89\% \\
        Quantum Volume & 1204 & 841 & 874 & \textbf{+27.41\%} & -3.92\% \\
        Random Avg & 208 & 208 & 163 & \textbf{+21.88\%} & \textbf{+21.79\%} \\
        \hline
    \end{tabular}
\end{subtable}
\hfill

\begin{subtable}[b]{0.48\textwidth}
    \caption{}
    \small 
    \setlength{\tabcolsep}{2.2pt}
    \centering
    \begin{tabular}{|lccccc|}
        \hline
         & \multicolumn{3}{c}{\textbf{Inter-core Comms.}} & \multicolumn{2}{c|}{\textbf{Improvement}} \\
         & \textbf{Russo} & \textbf{HQA} & \textbf{SQARL} & \textbf{vs Russo} & \textbf{vs HQA} \\\hline
        Cuccaro Adder & 517 & 110 & 74 & \textbf{+85.69\%} & \textbf{+32.73\%} \\
        Deutsch Jozsa & 363 & 36 & 36 & \textbf{+90.08\%} & \textbf{+0.00\%} \\
        Draper Adder & 2631 & 955 & 841 & \textbf{+68.03\%} & \textbf{+11.94\%} \\
        Graph State & 2892 & 1485 & 1990 & \textbf{+31.19\%} & -34.01\% \\
        QFT & 4887 & 1149 & 1586 & \textbf{+67.55\%} & -38.03\% \\
        QNN & 3549 & 2116 & 3349 & \textbf{+5.64\%} & -58.27\% \\
        Quantum Volume & 5289 & 3908 & 4241 & \textbf{+19.81\%} & -8.52\% \\
        Random Avg & 500 & 361 & 273 & \textbf{+45.46\%} & \textbf{+24.43\%} \\

        \hline
    \end{tabular}
\end{subtable}
\caption{Performance comparison with SOTA.
Results show the normalized allocation cost and cost improvement of our method with
RL state of the art (Russo) and with the overall state of the art (HQA) for circuits with
(a) 50 qubits and
(b) 100 qubits.
}
\label{tab:perf_comp_sota}
\end{table}

\section{Discussion}
\label{sec:discussion}

Tab.~\ref{tab:cost_my_versions} indicates that no allocation mode consistently outperforms the other; sequential seems to perform better on structured circuits such as the Draper Adder, QFT, or QNN, whereas parallel outperforms on less structured circuits, such as the set of random circuits.
It is uncertain why the parallel allocation mode cannot outperform sequential allocation across all circuits; one potential reason is that training the model exclusively in sequential mode hinders parallel allocation in some scenarios.

Regarding the inter-core communication cost benchmark shown in Tab.~\ref{tab:perf_comp_sota}, SQARL consistently outperforms the previous RL state-of-the-art (Russo) across almost all benchmarks.
The cost reduction is significant, particularly for structured circuits such as the Cuccaro Adder, Deutsch-Jozsa, Draper Adder, and QFT.
We also match or outperform HQA for half of the circuit types, particularly for random circuits.
For medium-sized circuits (50 qubits), our method remains notably close to HQA.
These results show that our RL approach can learn policies that approach the efficiency of hand-crafted heuristics.
In random circuits, SQARL outperforms the overall state of the art with a cost reduction of 21.90\% for 50 qubits and 24.48\% for 100 qubits.
Regarding scalability, as the problem size doubles from 50 to 100 qubits and the core count doubles, the performance gap between our method and Russo’s widens.
This is notable, as the maximum number of qubits during training is 20, which highlights that, in addition to being flexible with respect to problem size, our method can maintain the best performance, even for circuit dimensions it has never encountered before.
QNN and Graph State appear to be the most challenging benchmarks.

The average advantage of 21.79\% and 24.43\% over the set of 64 random circuits shows a solid performance gain in these types of circuits.
Although they do not form a specific category of circuits (they represent samples from the space of all possible circuits), the vast majority are much less structured than the quantum circuits in the literature.
The latter type of circuits frequently exhibits specific gate patterns and sequence repetition that seldom appear in the former.
Thus, RL algorithms rarely have a chance to train on these types of circuits, which dominate real-world applications of quantum computing.
If more ``realistic'' circuit generation techniques were available, RL methods could potentially outperform non-learning ones in the other types of circuits as well.

\section{Conclusion and Future Work}
\label{sec:conclusion}

RL methods for qubit allocation lagged behind the overall state of the art in terms of allocation cost and flexibility to the number of qubits and core interconnection topology, as discussed in Sec.~\ref{sec:related}. 
However, the results from Sec.~\ref{sec:results} show consistent improvement over the RL state of the art in allocation cost and highly competitive overall performance, with inter-core communication costs that improved or matched HQA's on half of the benchmark circuits.
Notably, in unstructured and random circuits, there is up to 25\% improvement in allocation cost with respect to the best method and up to 46\% with the previous best RL algorithm.
In addition, the policy is hardware-agnostic and does not require retraining if the quantum hardware changes.
The results from Sec.~\ref{sec:results} indicate that future iterations of SQARL (or other RL techniques) could push the allocation cost below that of HQA for the remaining quantum circuits.

Future work could focus on improved circuit generation for training.
The performance improvement of SQARL is most notable in random circuits, the type of circuit with which the policy was trained.
Although it represents a uniform distribution over all possible circuits, the results indicate that real circuits likely lie in a specific subspace rarely sampled during training.
Another aspect is that of accelerating the allocation process.
Most qubits do not change core from one slice to the other.
A system that detects which qubits require reallocation and executes the policy exclusively on those qubits could substantially reduce execution times.
Finally, circuit encoding relied on a set of hand-crafted features.
Methods like GNNs or topological ML could be leveraged to extract sets of learned circuit features that improve upon those proposed in Sec.~\ref{subsec:pol_model}.

\section*{Code availability}
\label{sec:code}

The complete implementation is publicly available at \url{https://github.com/Vicara12/SQARL}, along with the benchmarking circuits in JSON format \cite{Vicara12_MAI_TFM_2023}.

\section*{CRediT authorship contribution statement}
\textbf{V\'ictor Carballo}: conceptualization, software, formal analysis, validation, investigation, visualization, methodology, writing - original draft, writing - review \& editing. \textbf{J\'ulia L\'opez-Closa}: conceptualization, investigation, methodology, writing - original draft, writing - review \& editing.
\textbf{Mario Martin}: conceptualization, resources, supervision, funding acquisition, project administration, writing - review \& editing.

\section*{Declaration of competing interest}

The authors declare that they have no known competing financial interests or personal relationships that could have appeared to influence the work reported in this paper.

\section*{Acknowledgments}
This work has been funded by the Ministry of Economic Affairs and Digital Transformation of the Spanish Government through the QUANTUM ENIA project call – Quantum Spain project, and by the European Union through the Recovery, Transformation and Resilience Plan – NextGenerationEU within the framework of the Digital Spain 2026 Agenda.

\bibliography{references}
\bibliographystyle{template}

\end{document}